\documentclass[a4paper, 10pt, conference]{ieeeconf}      % Use this line for a4 paper

\IEEEoverridecommandlockouts                              % This command is only needed if
\usepackage{cite}
\usepackage{amsmath,amssymb,amsfonts}
\usepackage{algpseudocode,algorithm}
\usepackage{subfigure}
\usepackage{graphicx}

\title{\LARGE \bf
Towards Deep Robot Learning with Optimizer applicable to Non-stationary Problems
}

\author{Taisuke Kobayashi$^{1}$% <-this % stops a space
\thanks{$^{1}$Taisuke Kobayashi is with the Division of Information Science, Nara Institute of Science and Technology, 8916-5 Takayama-cho, Ikoma, Nara 630-0192, Japan
        {\tt\footnotesize kobayashi@is.naist.jp}}%
}

\begin{document}

\maketitle
\thispagestyle{empty}
\pagestyle{empty}

%%%%%%%%%%%%%%%%%%%%%%%%%%%%%%%%%%%%%%%%%%%%%%%%%%%%%%%%%%%%%%%%%%%%%%%%%%%%%%%%
\begin{abstract}

This paper proposes a new optimizer for deep learning, named d-AmsGrad.
In the real-world data, noise and outliers cannot be excluded from dataset to be used for learning robot skills.
This problem is especially striking for robots that learn by collecting data in real time, which cannot be sorted manually.
Several noise-robust optimizers have therefore been developed to resolve this problem, and one of them, named AmsGrad, which is a variant of Adam optimizer, has a proof of its convergence.
However, in practice, it does not improve learning performance in robotics scenarios.
This reason is hypothesized that most of robot learning problems are non-stationary, but AmsGrad assumes the maximum second momentum during learning to be stationarily given.
In order to adapt to the non-stationary problems, an improved version, which slowly decays the maximum second momentum, is proposed.
The proposed optimizer has the same capability of reaching the global optimum as baselines, and its performance outperformed that of the baselines in robotics problems.

\end{abstract}

%%%%%%%%%%%%%%%%%%%%%%%%%%%%%%%%%%%%%%%%%%%%%%%%%%%%%%%%%%%%%%%%%%%%%%%%%%%%%%%%
\section{Introduction}

The remarkable development of deep learning (DL)~\cite{lecun2015deep} has led to a number of technological innovations in the field of robotics that use deep learning in various aspects.
 DL is mainly used for robot vision~\cite{shvets2018automatic}, but recently, the robot control field also utilizes it for such as controller in combination with reinforcement learning~\cite{tsurumine2019deep} and; dynamics model~\cite{kobayashi2020q}.

However, utilizing DL technologies, which have been developed mainly for large-scale image classification~\cite{krizhevsky2012imagenet}, as tools in a straightforward manner would reveal the gaps with the problems in the robotics field.
That is, it is desirable for autonomous robots to set up problems with as little human intervention as possible, such as annotation and outlier removal, thereby getting the dataset with noise and outliers that cannot be ignored.
To find the optimal solution hidden in such noisy dataset, we have to newly develop DL technologies suitable for robotics.

This paper, therefore, focuses on stochastic gradient decent (SGD) based optimizers~\cite{robbins1951stochastic,kingma2014adam,reddi2019convergence,ilboudo2020tadam} as one of the fundamental DL technologies.
On the one hand, in our previous work~\cite{ilboudo2020tadam}, the first momentum of gradients, which yields the smooth update and is employed in the latest optimizers like Adam~\cite{kingma2014adam}, was re-derived based on student-t distribution to achieve robustness of noise and outliers.
As a result, it significantly improved the learning performances on the noisy dataset.
On the other hand, AmsGrad~\cite{reddi2019convergence} has been proposed to modify the second momentum of gradients, which adjusts learning rate for each parameter, in order to emphasize the update effect of the dominant (i.e., maximum) gradient while proving its convergence.
However, we can find that AmsGrad is hardly used in robot applications~\cite{hristov2020disentangled}, and furthermore, it often does not improve the learning performance compared to that of its original (i.e., Adam~\cite{reddi2019convergence})~\cite{ficht2018nimbro}.

\begin{figure}[tb]
    \centering
    \includegraphics[keepaspectratio=true,width=0.85\linewidth]{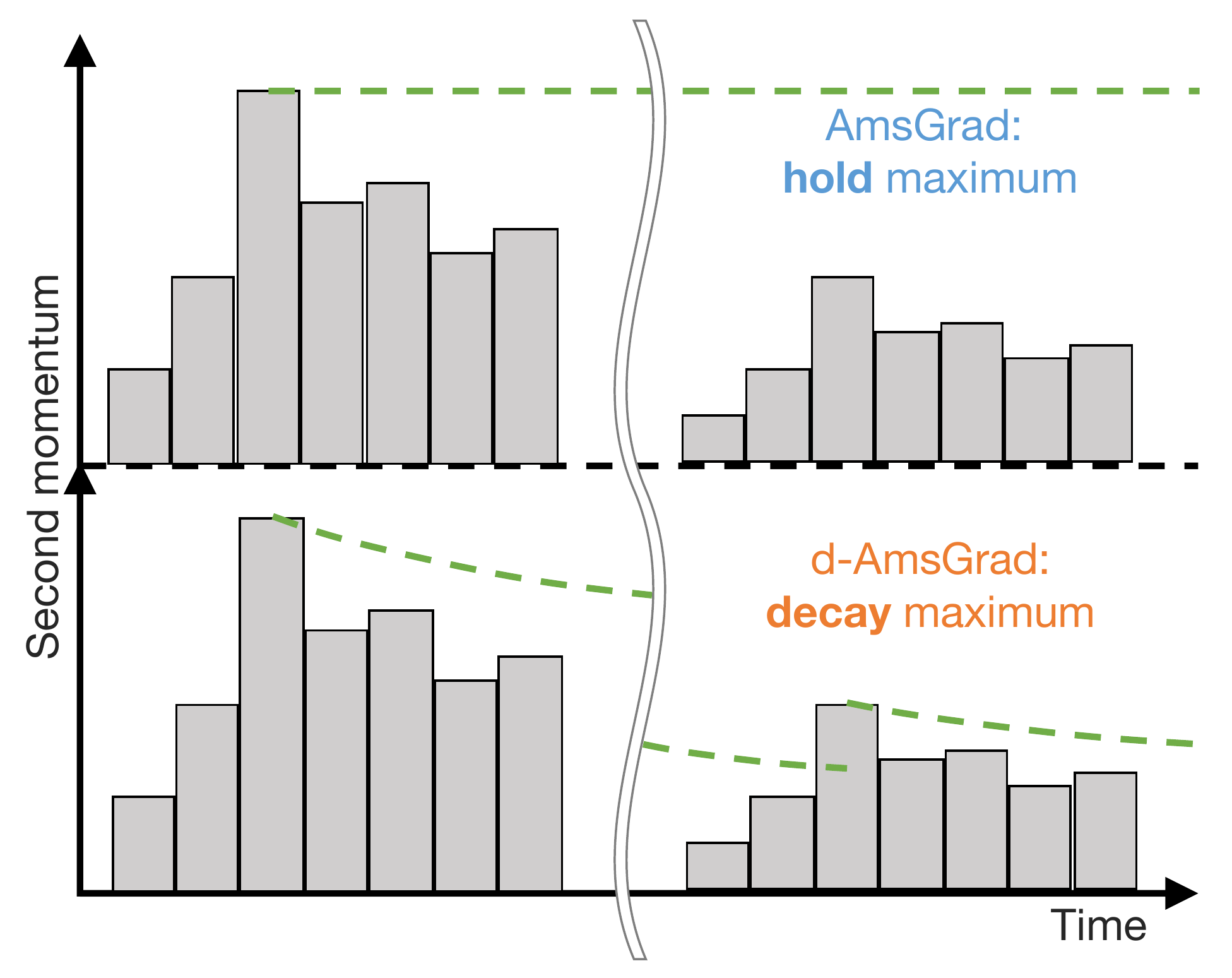}
    \caption{Second momentum on Non-stationary problem:
        as illustrated on the top, AmsGrad~\cite{reddi2019convergence} cannot follow the change of problem;
        in contrast, the proposed d-AmsGrad on the bottom can achieve adaptive learning.
    }
    \label{fig:concept}
\end{figure}

The reason why AmsGrad does not work well is considered that most of robot learning problems are non-stationary, but AmsGrad assumes the maximum gradient during learning to be stationarily given (see the top of Fig.~\ref{fig:concept}).
This gap would lose adaptability of learning rate whereas not emphasizing the maximum gradient correctly.
If AmsGrad is not used, we can ignore this gap problem, but in that case, we have to sacrifice the benefits of AmsGrad.

Hence, this paper proposes a new variant of AmsGrad to adapt to the non-stationary problems, named d-AmsGrad.
Although AmsGrad stores the maximum second momentum as it is, the proposed optimizer decays it slowly.
By doing so, even if the tendency of gradients is changed as time goes on, the proposed optimizer can find the new maximum second momentum while getting the benefits of AmsGrad in short term (see the bottom of Fig.~\ref{fig:concept}).

%Figure

To verify the proposed optimizer, d-AmsGrad, the following experiments are conducted.
Through a non-convex function benchmark, the same convergence performance as the baselines is illustrated.
As non-stationary learning problems, two bootstrapped learning methods are combined with the proposed and conventional optimizers.
First method is reinforcement learning~\cite{sutton2018reinforcement}, and it is applied to two benchmark tasks simulated by OpenAI Gym~\cite{brockman2016openai} with PyBullet~\cite{coumans2016pybullet}.
Second method is for learning latent dynamics~\cite{kobayashi2020q}, and it is applied to the prediction task of hexapod walking motion.
In these two experiments, the proposed optimizer outperformed the conventional ones.

%%%%%%%%%%%%%%%%%%%%%%%%%%%%%%%%%%%%%%%%%%%%%%%%%%%%%%%%%%%%%%%%%%%%%%%%%%%%%%%%
\section{Preliminaries}

%%%%%%%%%%%%%%%%%%%%%%%%%%%%%%%%%%%%%%%%
\subsection{Deep learning}

DL is the methodology to approximate blackbox functions by multi-layered neural networks~\cite{lecun2015deep}.
When $L$ number of hidden layers are given, the parameter set of that networks is basically summarized as $\theta = [w_1, b_1, \ldots, w_L, b_L, w_o, b_o]$ with weights $w_i$ and biases $b_i$ ($i = 1, \ldots, L$) for the respective hidden layers and $w_o$ and $b_o$ for the output layer.

Given $f_i(\cdot)$ as the nonlinear activation functions (usually common for all the layers), the outputs from the respective hidden layers $h_i$ are defined as follows:
\begin{align}
    h_i =
    \begin{cases}
        f_i(w_i^\top x + b_i) & i = 1
        \\
        f_i(w_i^\top h_{i-1} + b_i) & i \neq 1
    \end{cases}
    \label{eq:deep_hidden}
\end{align}
where $x$ denotes the input to the networks.
Next, the output from the last hidden layer $x_L$ is mapped to the domain for the learning target $y$.
\begin{align}
    y = g(w_o^\top x_L + b_o)
    \label{eq:deep_output}
\end{align}
where $g(\cdot)$ denotes the mapping function, such as a sigmoid function for $[0, 1]$.

With the initial parameter set $\theta_0$, which is randomly given, $y$ is naturally different from the desired output.
This difference is defined as the loss function $\mathcal{L}(x; \theta)$, and the loss for each data is computed as a scalar value.
For example, in the case of a regression problem, the following mean squared error (MSE), $\| d(x) - y(x; \theta) \|^2 / 2$, is commonly used.
Here, $d(x)$ denotes the desired output for $x$.

%%%%%%%%%%%%%%%%%%%%%%%%%%%%%%%%%%%%%%%%
\subsection{Stochastic gradient descent and momentum utilization}

To minimize the loss function like MSE, $\theta$ is updated according to the backpropagation of the loss function and its stochastic gradient descent (SGD) optimizer~\cite{robbins1951stochastic}.
That is, at time $t$, the gradient for each parameter is first computed as follows:
\begin{align}
    g_t = \nabla_\theta \mathcal{L}(x_t; \theta_{t-1})
    \label{eq:loss_grad}
\end{align}
Note that $x_t$ can be mini-batch dataset.
This gradient has the information about the direction in which the parameter should be changed to reduce the loss.
SGD therefore updates $\theta$ simply as follows:
\begin{align}
    \theta_t = \theta_{t-1} - \alpha g_t
    \label{eq:sgd}
\end{align}
where $\alpha$ denotes the learning rate.

Although $\theta$ can be updated according to eq.~\eqref{eq:sgd}, it has some numerical problems, such as oscillatory behavior and difficulty in tuning the learning rate.
In the latest improvements for SGD like Adam~\cite{kingma2014adam}, the first and second momenta of the gradient are employed.
Specifically, with two decaying factors $\beta_1, \beta_2$, the new update rule is given as follows:
\begin{align}
    m_t &= \beta_1 m_{t-1} + (1 - \beta_1) g_t
    \label{eq:momentum_1} \\
    v_t &= \beta_2 v_{t-1} + (1 - \beta_2) g_t^2
    \label{eq:momentum_2} \\
    \theta_t &= \theta_{t-1} - \frac{\alpha}{\sqrt{v_t (1 - \beta_2^t)^{-1}} + \epsilon} \frac{m_t}{1 - \beta_1^t}
    \label{eq:adam}
\end{align}
where the minuscule value $\epsilon$ is given for avoiding zero division.
That is, the former in the second term of eq.~\eqref{eq:adam} corresponds to the adaptive learning rate and the latter denotes the gradient from which the oscillatory component is removed.
In this way, the introduction of the first and second momenta in eqs.~\eqref{eq:momentum_1} and~\eqref{eq:momentum_2} significantly improves DL.

%%%%%%%%%%%%%%%%%%%%%%%%%%%%%%%%%%%%%%%%%%%%%%%%%%%%%%%%%%%%%%%%%%%%%%%%%%%%%%%%
\section{AmsGrad with decaying maximum momentum: d-AmsGrad}

%%%%%%%%%%%%%%%%%%%%%%%%%%%%%%%%%%%%%%%%
\subsection{AmsGrad and its problem}

For further improvement of the above update rule with the proof of convergence, AmsGrad has been proposed~\cite{reddi2019convergence}.
In this optimizer, the maximum value of the second momentum (let's call the second maximum momentum $v^\mathrm{max}$) is utilized to properly reflect the effects of large gradients in $\theta$.
That is, eq.~\eqref{eq:adam} is modified as follows:
\begin{align}
    v_t^\mathrm{max} &= \max(v_{t-1}^\mathrm{max}, v_t)
    \label{eq:momentum_max} \\
    \theta_t &= \theta_{t-1} - \frac{\alpha}{\sqrt{v_t^\mathrm{max} (1 - \beta_2^t)^{-1}} + \epsilon} \frac{m_t}{1 - \beta_1^t}
    \label{eq:amsgrad}
\end{align}
Thanks to the introduction of $v^\mathrm{max}$, it is now possible to optimize $\theta$ based on the gradients of high importance while ignoring the small gradients.
In addition, AmsGrad can be integrated into the other optimizers that use the second momentum.

Despite these advantages of AmsGrad, it has been reported that AmsGrad is rarely used in the DL for robotics field~\cite{hristov2020disentangled} and it often has no improvement~\cite{ficht2018nimbro}.
We can find that the reason for this gap between theory and reality comes from the non-stationarity of real robotic problems.
That is, the desired output $d(x)$ is assumed, and it is stationarily given in supervised learning.
However, in the robotic problems, $d(x)$ will be changed as time goes on.
For example, for robot control using deep reinforcement learning~\cite{tsurumine2019deep}, no true $d(x)$ is given and it should be estimated by the same or another neural networks since this problem is a kind of bootstrapped learning.
Alternatively, in multi-task learning~\cite{rahmatizadeh2018vision}, it is easy to imagine that the scale of the gradients varies depending on the tasks.

%%%%%%%%%%%%%%%%%%%%%%%%%%%%%%%%%%%%%%%%
\subsection{Proposal}

In this section, we aim to apply AmsGrad to the problems where the supervised signals are non-stationary, without sacrificing its benefits.
The simple but practical solution for this is to assume that the value of the second maximum momentum decays slowly as its information will drift away from the real situation.
Even if the problem to be solved is non-stationary, this decaying allows the optimizer to follow the new situation; and even otherwise, if the decaying speed is enough slower than the frequency of occurrence of the maximum gradient, the adverse effects of decaying would be negligible.

Specifically, this concept is implemented as a new optimizer, named d-AmsGrad.
Its difference from the original AmsGrad is only in eq.~\eqref{eq:momentum_max} as follows:
\begin{align}
    v_t^\mathrm{max} &= \max(\beta_3 v_{t-1}^\mathrm{max}, v_t)
    \label{eq:momentum_dmax}
\end{align}
where $\beta_3$ denotes the additional decaying factor.
The overall implementation integrated with Adam is shown in Alg.~\ref{alg:proposal}.

%Algorithm
\begin{algorithm}[tb]
    \caption{d-AmsGrad optimizer}
    \label{alg:proposal}
    \begin{algorithmic}[1]
        % requirements
        \Require{$\alpha$: Learning rate}
        \Require{$\beta_1$, $\beta_2 \in$ [0,1): Decaying factors of momenta}
        \Require{$\epsilon$: Minuscule value for numerical stabilization}
        \Require{$\beta_3 \in [\beta_2, 1]$: Additional decaying factor}
        \Require{$\mathcal{L}(x_t; \theta)$: Loss function with parameter set $\theta$}
        % initialization
        \State{$\theta_0 \gets$ random, $m_0 \gets 0$, $v_0 \gets 0$, $v_0^\mathrm{max} \gets 0$, $t \gets 0$}
        % main loop
        \While{$t<T_\mathrm{max}$}
            \State{$t \gets t+1$}
            \State{$x_t \gets x_{t+1}$}
            \State{$g_t \gets \nabla_{\theta}\mathcal{L}(x_t; \theta_{t-1})$}
            \State{$m_t \gets \beta_1 m_{t-1} + (1 - \beta_1) g_t$}
            \State{$v_t \gets \beta_2 v_{t-1} + (1 - \beta_2) g_t^2$}
            \State{$v_t^\mathrm{max} \gets \max(\beta_3 v_{t-1}^\mathrm{max}, v_t)$}
            \Comment{New implementation}
            \State{$\theta_t \gets \theta_{t-1} - \alpha \frac{m_t}{(1 - {\beta}^t_1)(\sqrt{v_t^\mathrm{max}(1 - {\beta}^t_2)^{-1}} + \epsilon)}$}
        \EndWhile
        % end
        \State{\Return{$\theta_t$}}
    \end{algorithmic}
\end{algorithm}

%%%%%%%%%%%%%%%%%%%%%%%%%%%%%%%%%%%%%%%%
\subsection{Analysis}

First of all, three modes can be expected according to $\beta_3$ in d-AmsGrad.
\begin{enumerate}
    \item $\beta_3 \leq \beta_2$:
    If the decaying speed of the second maximum momentum is faster than that of the second momentum, eq.~\eqref{eq:momentum_dmax} is always $v_t^\mathrm{max} = v_t$ since $g_t^2 > 0$ is added to $v_t$.
    That is, d-AmsGrad returns to the optimizer without it.
    \item $\beta_3 = 1$:
    Since eq.~\eqref{eq:momentum_dmax} perfectly matches eq.~\eqref{eq:momentum_max}, d-AmsGrad returns to AmsGrad.
    \item $\beta_3 \in (\beta_2, 1)$:
    With the slower decaying by $\beta_3$ than that by $\beta_2$, the desired adaptive behavior can be expected.
\end{enumerate}
From these three modes, we can agree that d-AmsGrad connects the original optimizer like Adam to that with AmsGrad continuously.

For the third mode, we consider when the max operation in eq.~\eqref{eq:momentum_dmax} replaces $v_t^\mathrm{max}$ to $v_t$.
Given the time when the last replacement occurred, $T$, the inequality for judging the replacement at $T+t$ is derived as follows:
\begin{align}
    \beta_3 v_{T+t-1}^\mathrm{max} &\leq \{ \beta_2 v_{T+t-1} + (1 - \beta_2) g_{T+t}^2 \}
    \nonumber \\
    \beta_3^t v_{T}^\mathrm{max} &\leq \left \{ \beta_2^t v_{T}^\mathrm{max} + (1 - \beta_2) \sum_{k=0}^{t-1} \beta_2^{k} g_{T+t-k}^2 \right \}
    \nonumber \\
    % (\beta_3^t - \beta_2^t) v_{T}^\mathrm{max} &\leq (1 - \beta_2) \sum_{k=0}^{t-1} \beta_2^{k} g_{T+t-k}^2
    % \nonumber \\
    v_{T}^\mathrm{max} &\leq \frac{1 - \beta_2}{\beta_3^t - \beta_2^t} \sum_{k=0}^{t-1} \beta_2^{k} g_{T+t-k}^2
\end{align}
Here, if the recent second momenta $g_{T+t-k}^2$ have a common expected value $\bar v_T$, the following inequality can be expected.
\begin{align}
    v_{T}^\mathrm{max} \leq \frac{1 - \beta_2}{\beta_3^t - \beta_2^t} \sum_{k=0}^{t-1} \beta_2^{k} \bar v_T
    = \frac{1 - \beta_2^t}{\beta_3^t - \beta_2^t} \bar v_T
    \label{eq:damsgrad_anal}
\end{align}
In the original AmsGrad (with $\beta_3 = 1$), the replacement is never performed unless $\bar v_T$ is directly over $v_{T}^\mathrm{max}$.
In contrast, since the coefficient of $\bar v_T$ in eq.~\eqref{eq:damsgrad_anal} increases over time (and becomes infinity) when $\beta_3 > \beta_2$, the replacement regularly occurs in d-AmsGrad, thereby allowing the optimizer following to new scales of the gradients.

%%%%%%%%%%%%%%%%%%%%%%%%%%%%%%%%%%%%%%%%%%%%%%%%%%%%%%%%%%%%%%%%%%%%%%%%%%%%%%%%
\section{Experiments}

%%%%%%%%%%%%%%%%%%%%%%%%%%%%%%%%%%%%%%%%
\subsection{Conditons}

To verify the proposed optimizer, d-AmsGrad, three kinds of experiments are conducted.
Except for the first benchmark, neural networks implemented by PyTorch~\cite{paszke2017automatic} are employed.
Each component is built with five fully-connected layers, each of which includes 100 neurons.
Layer normalization~\cite{ba2016layer} and Swish function~\cite{elfwing2018sigmoid} are employed as the activation function of each hidden layer.

As a base optimizer, t-Adam~\cite{ilboudo2020tadam} is used.
Three hyperparameters are commonly set to the default values: $\beta_1 = 0.9$, $\beta_2 = 0.999$, and $\epsilon = 10^{-8}$.
Except for the first benchmark with the fine tuning of learning rate $\alpha$ and an additional hyperparameter for t-Adam (i.e., $k_\mathrm{dof}$), they are given to be $\alpha = 3 \times 10^{-4}$ and $k_\mathrm{dof} = 1$.
Using these configurations, three optimizers, i.e., t-Adam ($\beta_3 = \beta_2$), t-AmsGrad ($\beta_3 = 1$), and td-AmsGrad ($\beta_3 = 0.99999$), are compared.

%%%%%%%%%%%%%%%%%%%%%%%%%%%%%%%%%%%%%%%%
\subsection{Non-convex function benchmark}

%Figure
\begin{figure*}[tb]
    \centering
    \subfigure[t-Adam]{
        \includegraphics[keepaspectratio=true,width=0.315\linewidth]{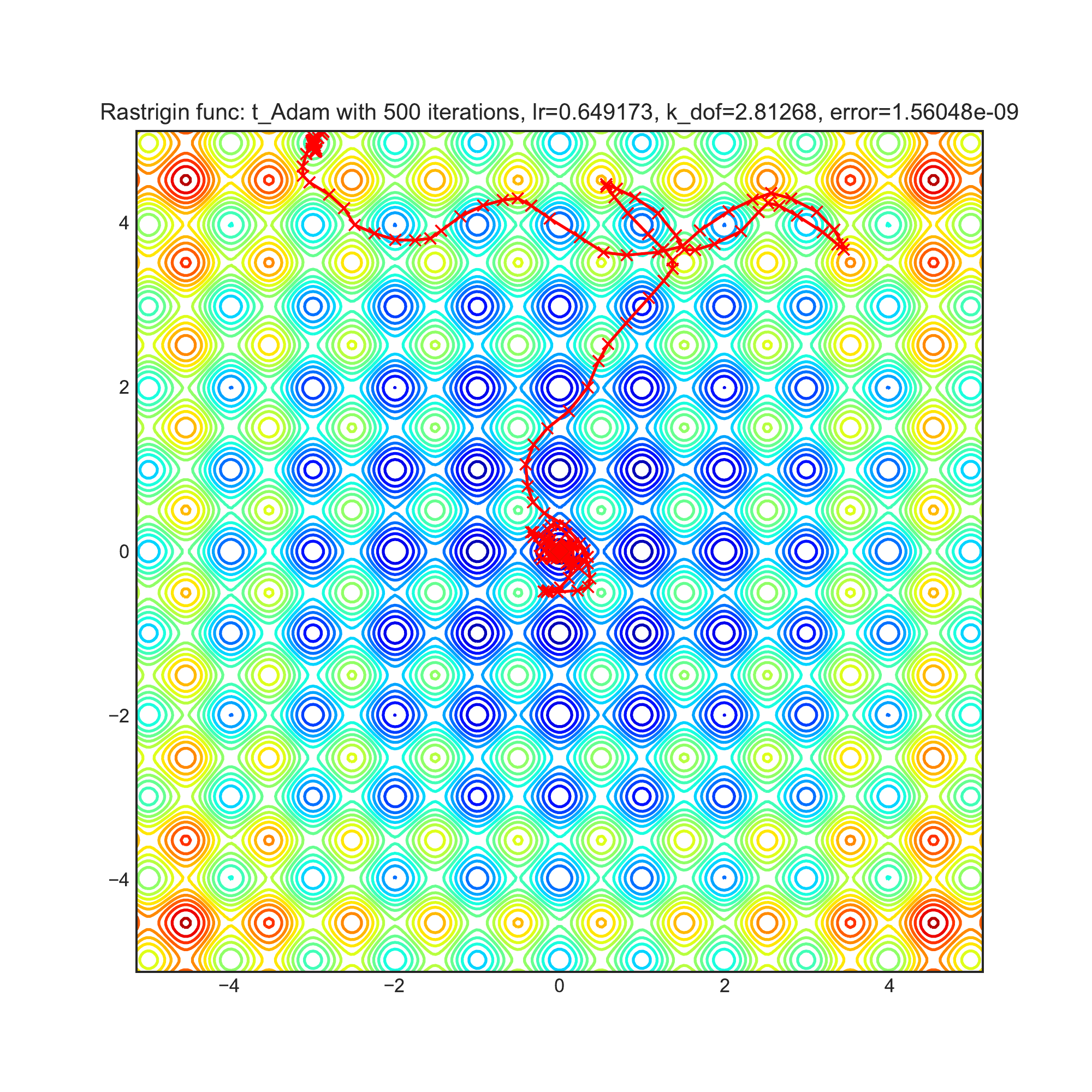}
    }
    \centering
    \subfigure[t-AmsGrad]{
        \includegraphics[keepaspectratio=true,width=0.315\linewidth]{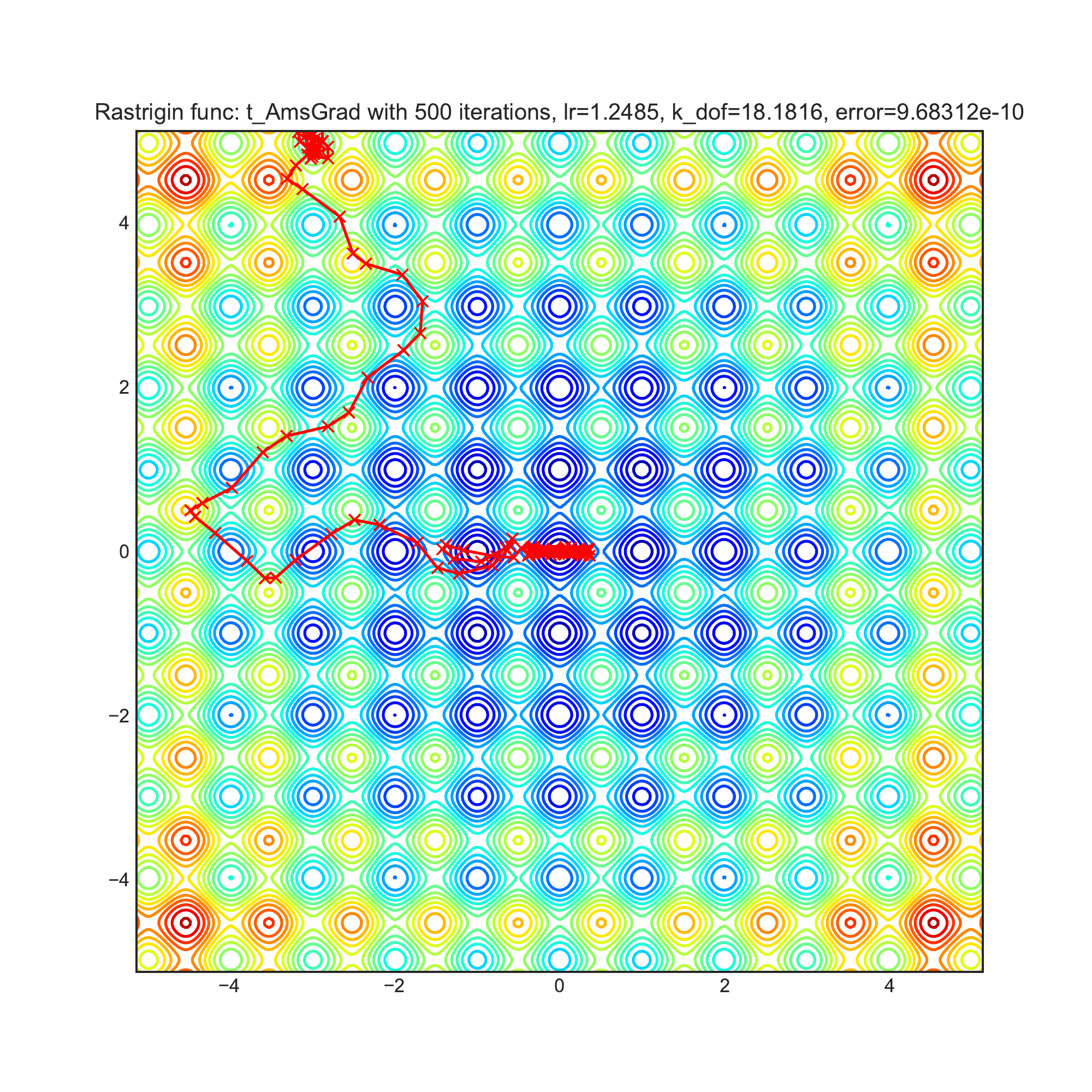}
    }
    \centering
    \subfigure[td-AmsGrad]{
        \includegraphics[keepaspectratio=true,width=0.315\linewidth]{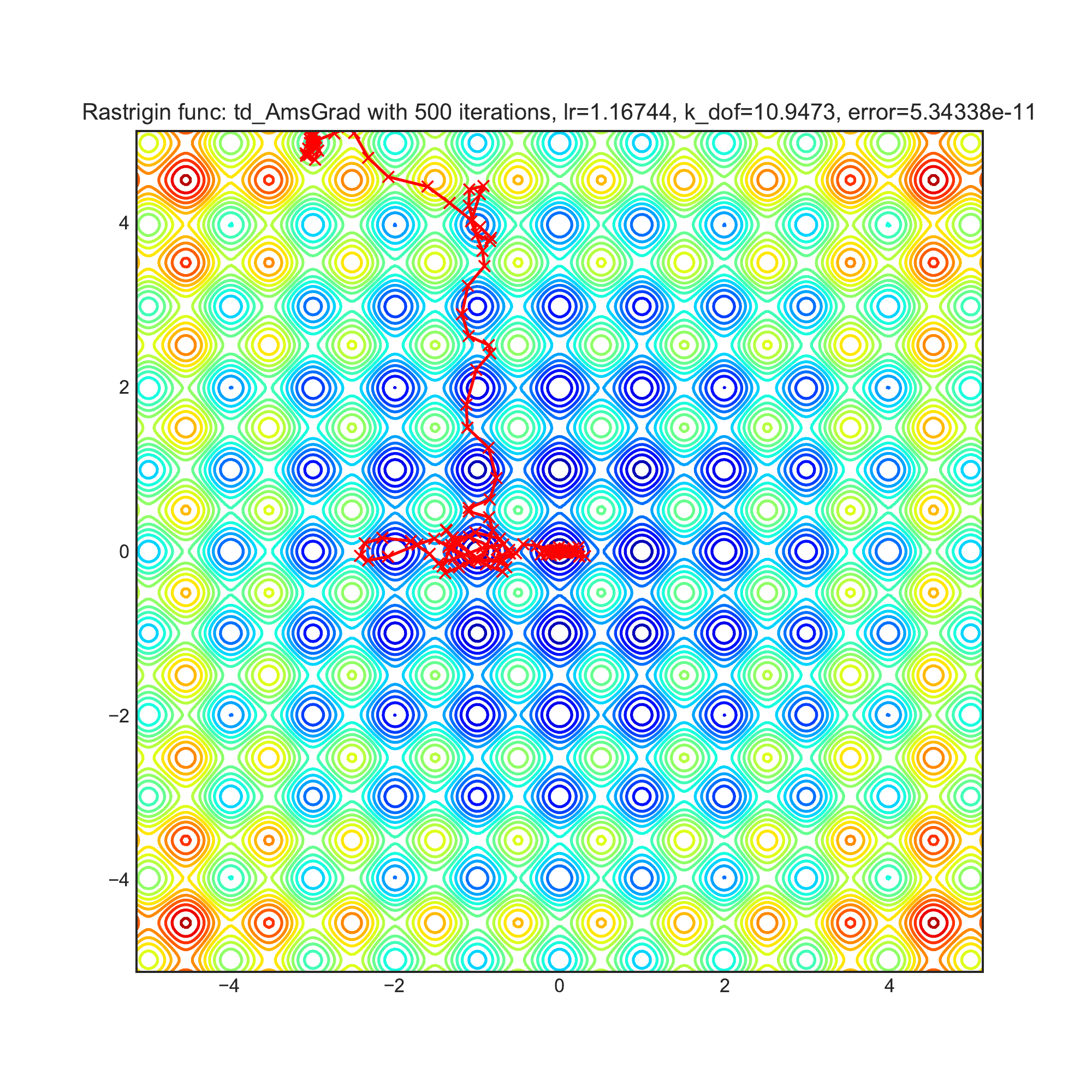}
    }
    \vspace{-4mm}
    \caption{Convergence performances on Rastrigin function benchmark:
        the initial point is $(-3, 5)$ and the global minimum is on the center $(0, 0)$;
        hyperparameters were optimized heuristically;
        the red lines illustrated the trajectories of parameters updates;
        although the point moved along different paths, all optimizers including the proposed method (c) eventually converged on the global minimum.
    }
    \label{fig:rastrigin}
\end{figure*}

First of all, to demonstrate the convergence capability of d-AmsGrad, a two-dimensional non-convex function (i.e., Rastrigin function) is solved.
\begin{align}
    \mathcal{L}(x_1, x_2) = 20 + \sum_{i=1}^2(x_i^2 - 10 \cos(2 \pi x_i))
    \label{eq:loss_rastrigin}
\end{align}
where $(x_1, x_2)$ are the parameters to be optimized by the three optimizers.
The initial point is set on $(-3, 5)$, and they move toward the center $(0, 0)$ while escaping a lot of local optima along that way.
Since the learning rate (and $k_\mathrm{dof}$) can be expected to be different from that for neural networks to solve real problems, they are heuristically optimized using Hyperopt~\cite{bergstra2015hyperopt}.

The results demonstrated by the three optimizers are illustrated in Fig.~\ref{fig:rastrigin}.
Although their trajectories were different from each other due to the different optimized hyperparameters, all of them successfully reached the global optimum.
The optimized learning rate for both t-AmsGrad and td-AmsGrad was higher than 1, compared to about 0.6 for t-Adam.
This is because the amount of updates is stable by utilizing the second (decaying) maximum momentum.
From these results, it can be seen that the proposed optimizer, d-AmsGrad, has the same converge capability as AmsGrad.

%%%%%%%%%%%%%%%%%%%%%%%%%%%%%%%%%%%%%%%%
\subsection{Reinforcement learning}

%Figure
\begin{figure*}[tb]
    \centering
    \subfigure[InvertedPendulum]{
        \includegraphics[keepaspectratio=true,width=0.315\linewidth]{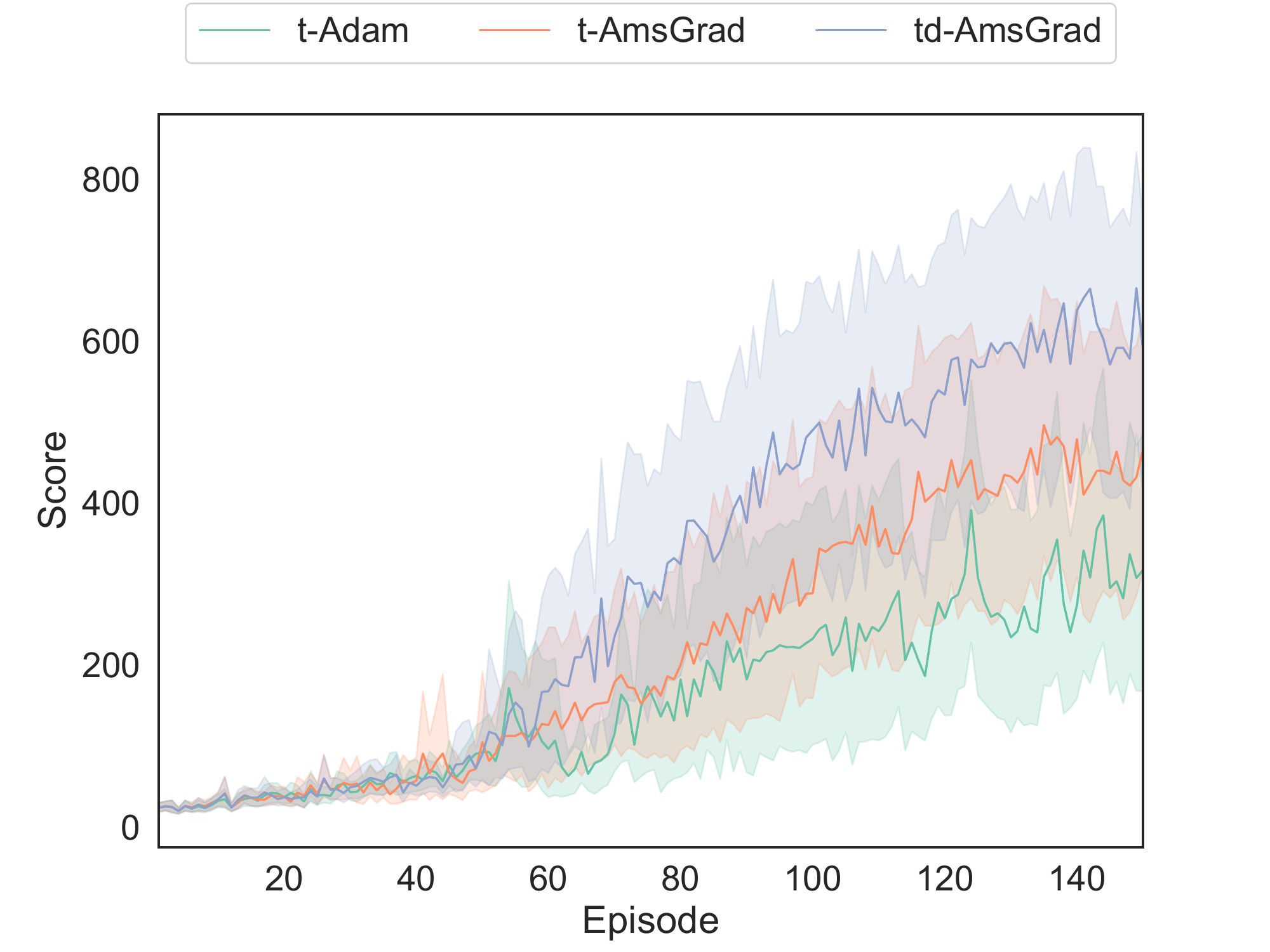}
    }
    \centering
    \subfigure[Swingup]{
        \includegraphics[keepaspectratio=true,width=0.315\linewidth]{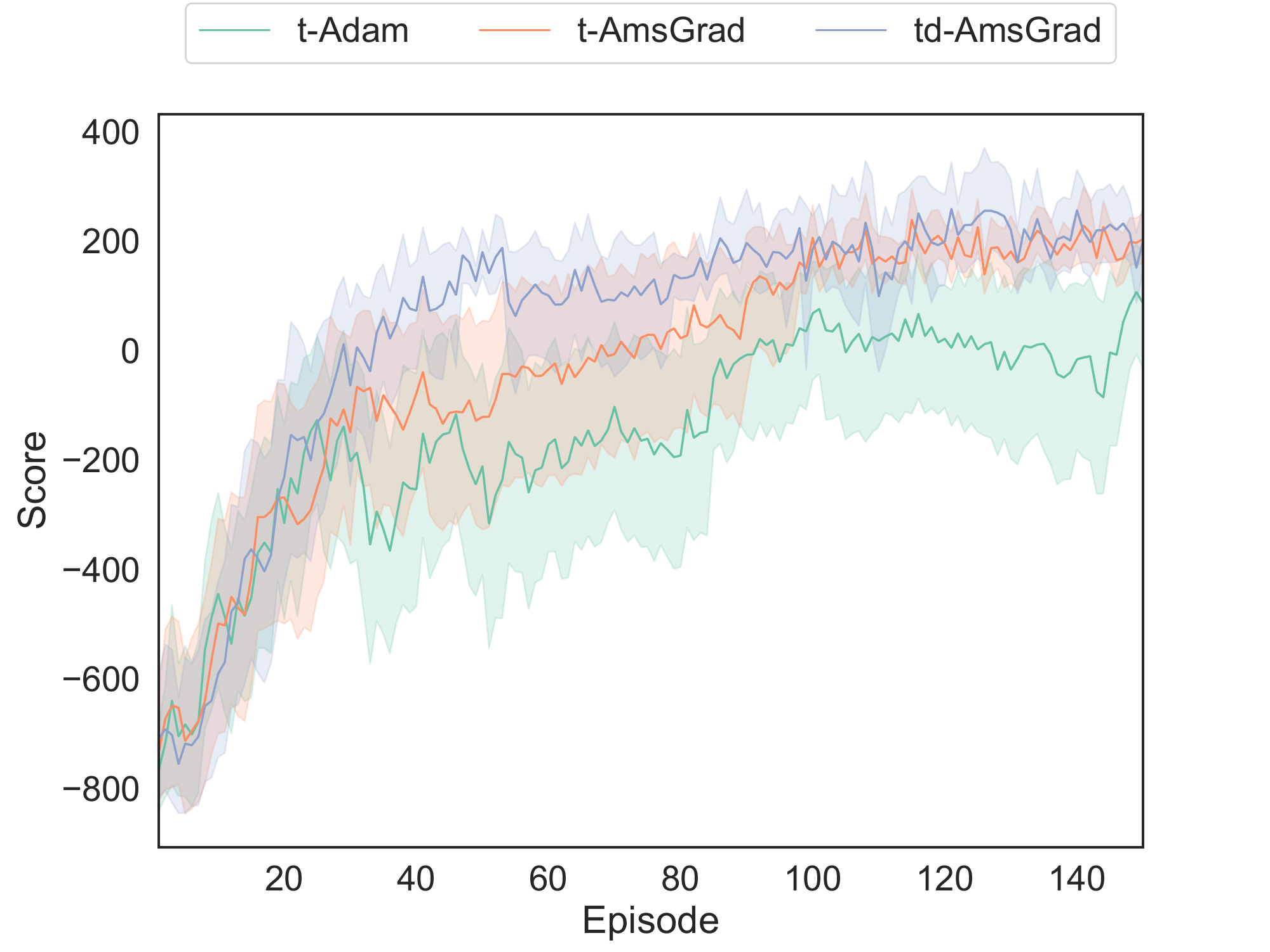}
    }
    \centering
    \subfigure[Summary]{
        \includegraphics[keepaspectratio=true,width=0.315\linewidth]{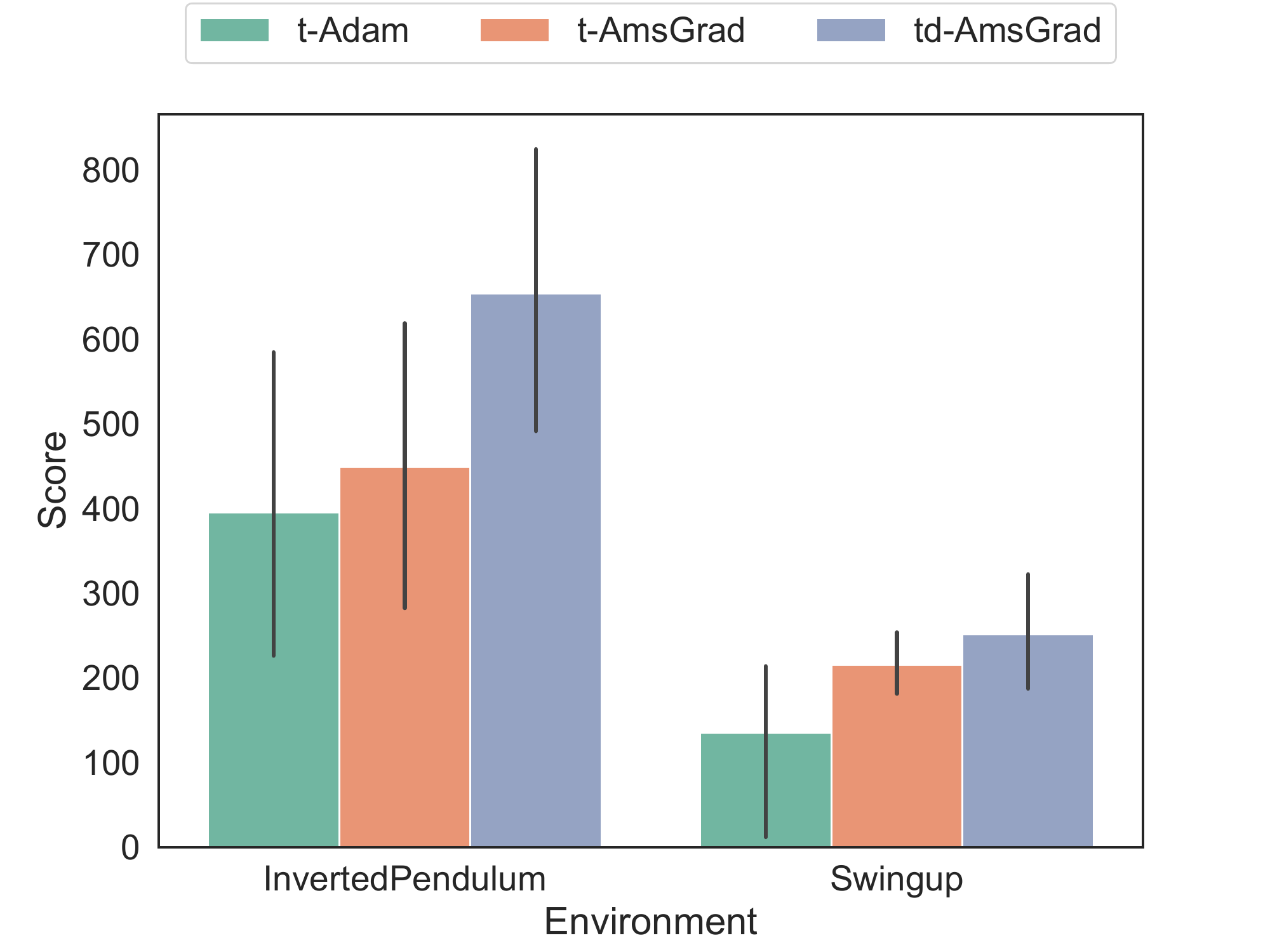}
    }
    \vspace{-4mm}
    \caption{Results of reinforcement learning benchmarks:
        for each optimizer, 20 trials with the different random seeds were performed;
        as shown in (a) and (b), the proposed method, td-AmsGrad gained the higher score (i.e. the sum of rewards in each episode) than that of the other optimizers;
        as a result, the td-AmsGrad outperformed the others in terms of the test score after learning.
    }
    \label{fig:result_rl}
\end{figure*}

As more realistic problems with non-stationarity, robot control problems in which the optimal control policy is learned by deep reinforcement learning is first conducted.
In general, reinforcement learning algorithms have no true signals about the policy, hence, they are implemented in a bootstrapped manner.
In this paper, an actor-critic algorithm with entropy and temporal difference regularizations~\cite{haarnoja2018soft,parisi2019td} is employed by minimizing the following loss function.
\begin{align}
    \mathcal{L}(s_t, a_t, s_{t+1}; \theta^{V,\pi}) &= - \delta(s_t, a_t, s_{t+1})
    \nonumber \\
    &\times (V(s_t; \theta^V) + \ln \pi(a_t \mid s_t; \theta^\pi))
    \label{eq:loss_rl}
\end{align}
where $s$ and $a$ denote state and action of the robot.
The temporal difference $\delta$ decides whether the sampled data is good or not.
Along with this direction, the value function $V$ and the stochastic policy $\pi$, which is parameterized as student-t distribution~\cite{kobayashi2019student} in this paper, are optimized.

Here, the learning results of two benchmark tasks, i.e., InvertedPendulum and Swingup implemented by OpenAI Gym~\cite{brockman2016openai} with Pybullet~\cite{coumans2016pybullet}, are depicted in Fig.~\ref{fig:result_rl}.
In both scenarios, t-Adam (without AmsGrad) could not acquire the given tasks well enough.
This is because the learning rate $3 \times 10^{-4}$ is relatively large for deep reinforcement learning, and therefore, t-Adam updated the value and policy functions unstably.
On the other hand, AmsGrad improved the performances (i.e., learning speed and final score) due to stable adaptation of learning rate.
In particular, td-AmsGrad made learning faster than t-AmsGrad, and that yielded the best performance in the three optimizers.
This result would be thanks to the adaptability to non-stationary problems, which enables the second maximum momentum to adjust for the most recent gradients.

%%%%%%%%%%%%%%%%%%%%%%%%%%%%%%%%%%%%%%%%
\subsection{Learning of latent dynamics}

%Figure
\begin{figure}[tb]
    \centering
    \includegraphics[keepaspectratio=true,width=0.9\linewidth]{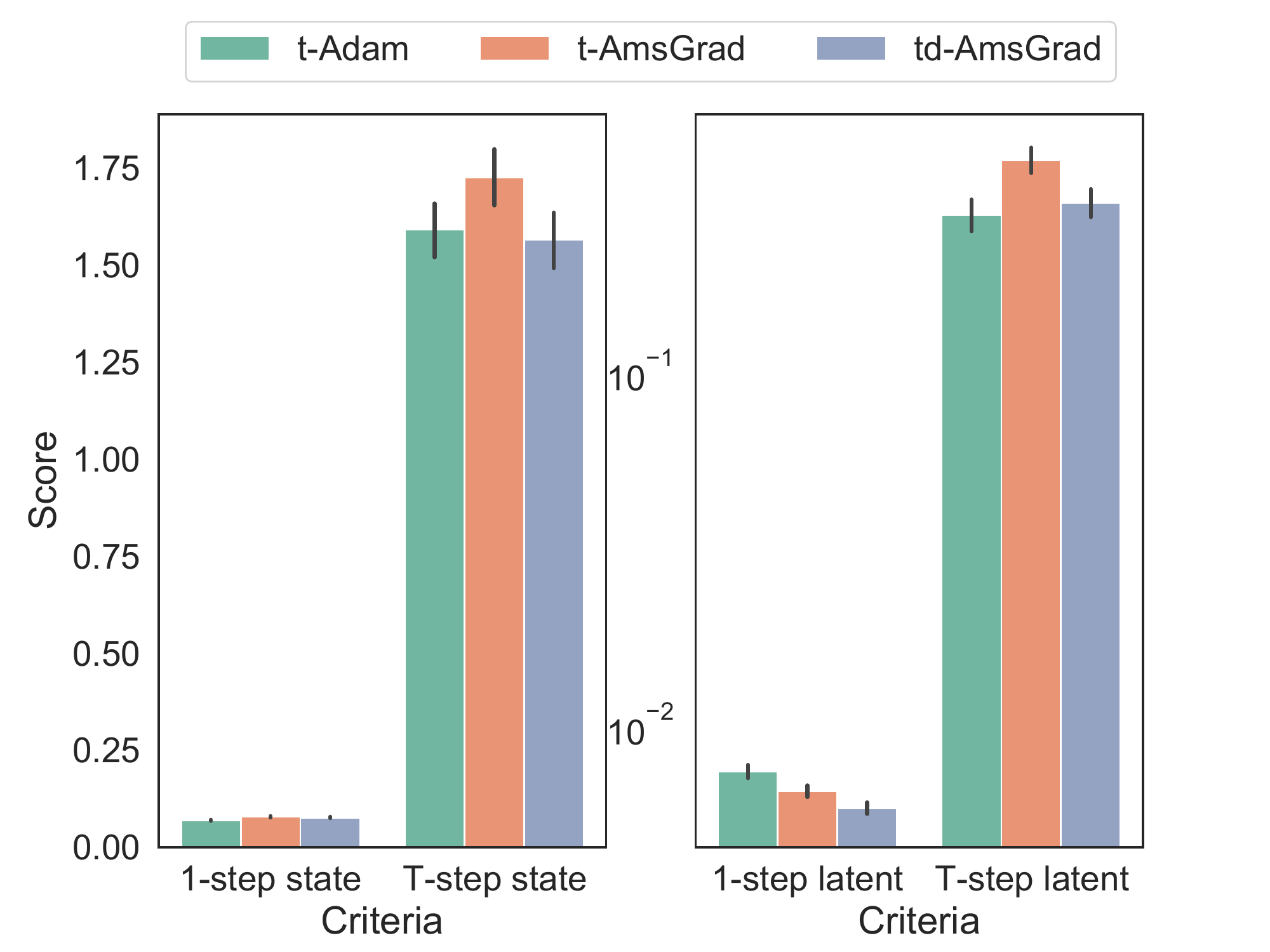}
    \vspace{-3mm}
    \caption{Prediction results of hexapod walking motions~\cite{kobayashi2020q}:
        for each optimizer, 20 trials with the different random seeds were performed;
        as for one-step latent prediction, the td-AmsGrad achieved the lower prediction error than the others;
        for T-step state and latent predictions, the proposed td-AmsGrad had the same-level prediction error as the t-Adam while the conventional t-AmsGrad increased its prediction error.
    }
    \label{fig:result_qvae_summary}
\end{figure}

As another learning problem with the bootstrapped learning method, motion prediction using latent dynamics is performed.
While the latent dynamics is with low computational cost for prediction and planning even in high-dimensional observation, it is basically learned in an unsupervised manner.
In q-VAE~\cite{kobayashi2020q}, three components, an encoder, the latent dynamics, and a decoder, are connected in series.
Its loss function is given as follows:
\begin{align}
    \mathcal{L}(x^\prime, x; \theta) &= - \ln_q{p(x^\prime \mid z^\prime(\theta^l) ; \theta^d)}
    \nonumber \\
    &+ \beta_q(x, z) \mathrm{KL}_q(\rho(z \mid x ; \theta^e) \mid \mid p(z))
    \nonumber \\
    &- \gamma \ln \rho(z^\prime(\theta^l) \mid x^\prime ; \theta^e)
\end{align}
where $z$ means the latent variable, and $\beta$, $q$, and $\gamma$ denote the hyperparameters for q-VAE (details are in the literature~\cite{kobayashi2020q}).
Although the last term is for training latent dynamics directly, the desired value is encoded from the next state, and is affected by the encoder's accuracy.

The dataset for predicting a hexapod walking motion generated by central pattern generators, which has been made in~\cite{kobayashi2020q}, is trained.
As a result, the prediction errors in state and latent spaces for the test data are summarized in Fig.~\ref{fig:result_qvae_summary}.
Here, ``1-step'' and ``T-step'' mean to predict all states from one step before and from the initial state, respectively.
As can be seen in the figure, although t-AmsGrad deteriorated the accuracy of T-step prediction, td-AmsGrad performed the same or better than t-Adam.
Unlike reinforcement learning, in this problem, the dataset itself is given in advance and new data cannot be searched additionally, and therefore, it is difficult to escape from the local optima if once the learning rate is decreased due to AmsGrad.
However, d-AmsGrad enabled the optimizer to recover the learning rate.

%%%%%%%%%%%%%%%%%%%%%%%%%%%%%%%%%%%%%%%%%%%%%%%%%%%%%%%%%%%%%%%%%%%%%%%%%%%%%%%%
\section{Conclusion}

In this paper, d-AmsGrad optimizer, which is the extended method of AmsGrad, was developed to adapt learning systems to non-stationary problems in the robotics field.
By introducing the slow decaying factor of the second maximum momentum, the proposed optimizer can follow even if the tendency of gradients is changed as time goes on.
In addition, thanks to it's slow decaying, it can keep the benefits of AmsGrad in short term.
The analysis of its behavior revealed its adaptability to the new expected second momentum.
In addition, d-AmsGrad matches the optimizer without AmsGrad if the decaying factor is smaller than the value for moving average of the second momentum, namely, the proposed optimizer connects the optimizer with/without AmsGrad continuously.
The convergence ability of d-AmsGrad in the non-convex function was demonstrated.
As the non-stationary problems, two types of the bootstrapped learning methods (i.e., reinforcement learning and learning of latent dynamics) are utilized in the respective problems.
In both problems, d-AmsGrad outperformed the conventional optimizers with/without AmsGrad.

Future work of this work is to develop an adaptive decaying method and to investigate the performance of d-AmsGrad on more realistic robot scenarios.

%%%%%%%%%%%%%%%%%%%%%%%%%%%%%%%%%%%%%%%%%%%%%%%%%%%%%%%%%%%%%%%%%%%%%%%%%%%%%%%%

% \section*{ACKNOWLEDGMENT}
%
% The preferred spelling of the word �acknowledgment� in America is without an �e� after the �g�. Avoid the stilted expression, �One of us (R. B. G.) thanks . . .�  Instead, try �R. B. G. thanks�. Put sponsor acknowledgments in the unnumbered footnote on the first page.
%
%
%
%%%%%%%%%%%%%%%%%%%%%%%%%%%%%%%%%%%%%%%%%%%%%%%%%%%%%%%%%%%%%%%%%%%%%%%%%%%%%%%%

%%%%%%%%%%%%%%%%%%%%%%%%%%%%%%%%%%%%%%%%%%%%%%%%%%%%%%%%%%%%%%%%%%%%%%%%%%%%%%%%
\bibliographystyle{IEEEtran}
{
\bibliography{bibliography}
}

\end{document}